\documentclass[10pt,twocolumn,letterpaper]{article}

\usepackage{cvpr}
\usepackage{times}
\usepackage{epsfig}
\usepackage{graphics}
\usepackage{graphicx}
\usepackage{amsmath}
\usepackage{amssymb}
\usepackage{amsfonts}

\usepackage[pagebackref=true,breaklinks=true,letterpaper=true,colorlinks,bookmarks=false]{hyperref}

\cvprfinalcopy 

\def\cvprPaperID{1045} 

\ifcvprfinal\fi
\begin{document}

\title{Personalized and Occupational-aware Age Progression by Generative Adversarial Networks}

\author{Siyu Zhou$^{~1}$~ Weiqiang Zhao$^{~1}$~ Jiashi Feng$^{~2}$~ Hanjiang Lai$^{~1}$~ Yan Pan$^{~1}$~ Jian Yin$^{~1}$~ Shuicheng Yan$^{~2,3}$\\
$^{1~}$Sun Yat-Sen University, China\\
$^{2~}$National University of Singapore, Singapore\\
$^{3~}$QiHoo 360 Artificial intelligence institute, China\\
}

\maketitle

\begin{abstract}
Face age progression, which aims to predict the future looks, is important for various applications and has been received considerable attentions. Existing methods and datasets are limited in exploring the effects of occupations which may influence the personal appearances. In this paper, we firstly introduce an occupational face aging dataset for studying the influences of occupations on the appearances. It includes five occupations, which enables the development of new algorithms for age progression and facilitate future researches. Second, we propose a new occupational-aware adversarial face aging network, which learns human aging process under different occupations. Two factors are taken into consideration in our aging process: personality-preserving and visually plausible texture change for different occupations. We propose personalized network with personalized loss in deep autoencoder network for keeping personalized facial characteristics, and occupational-aware adversarial network with occupational-aware adversarial loss for obtaining more realistic texture changes. Experimental results well demonstrate the advantages of the proposed method by comparing with other state-of-the-arts age progression methods.
\end{abstract}

\section{Introduction}

Face age progression~\cite{ramanathan2009age,fu2010age}, also called face aging, is to predict the future looks of a person. It is one of the key techniques for a variety of applications, including looking for the missing person, cross-age face analysis~\cite{park2010age} and so on. Recently, many research efforts have been devoted to generating realistic aged faces, which can be roughly divided into two categories: physical model-based age progression~\cite{suo2012concatenational,suo2010compositional} and prototype-based age progression~\cite{burt1995perception,kemelmacher2014illumination}. The physical model-based methods model the facial patterns and physical mechanisms of aging. While, the prototype-based age progression methods transfer the differences between prototypes (e.g., the average face of each group) into the individual faces. Deep learning methods have also been applied in face age progression due to their powerful feature representations. Wang et al.~\cite{wang2016recurrent} showed a recurrent face aging framework and Zhang et al.~\cite{zhang2017age} proposed conditional adversarial autoencoder framework (CAAE) for age progression.

\begin{figure*}[ht]
\begin{center}
\includegraphics[width=0.9\linewidth]
                   {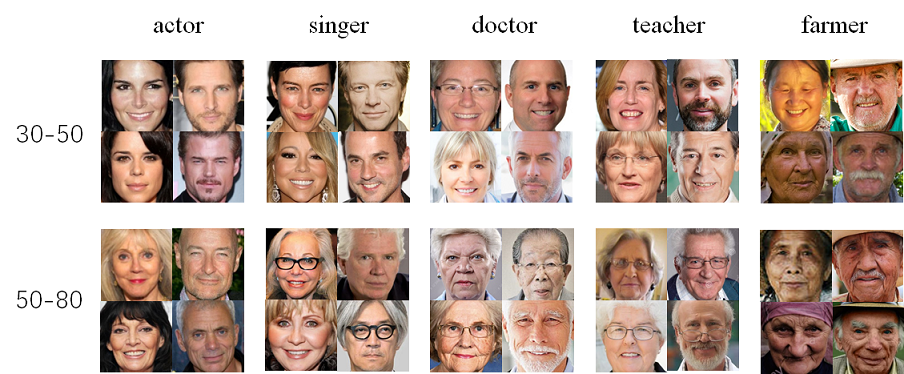}
\end{center}
   \caption{Examples of OFAD. According to different occupational information, we collect different images in two age groups (30-50, 50-80). The occupation from left to right is actor, singer, doctor, teacher and farmer. From left to right, we found that deeper and deeper the facial wrinkles and skin colors are. This contrast is more obvious in the old age groups, because the face has been affected by working environment longer.}
\label{fig:OFAD}
\end{figure*}

However, existing works always generate only one future look for a person, while totally ignore the future look of a person may change in different occupations. For example, a 20-years-old young man chooses actor/farmer as his career. When he is 50-years-old, the look of the farmer and the look of the actor may have some differences even for the same person. Different occupations may have different appearances~\cite{little2012evolution} as shown in Figure~\ref{fig:OFAD}. In this paper, we explore the impact of occupations on the age progression. Note that, for different occupations, the most perceptible difference is skin texture. Hence, we only focus on skin aging~\footnote{The face aging process can be divided into two stages~\cite{suo2010compositional,fu2010age}: child growth and adult aging. Shape change is the most prominent factor during the child growth, while the most greatest change is skin aging (texture change) during adult aging. We only focus on adult aging.} in this work.

We firstly introduce \textit{occupational face aging dataset} (OFAD). Figure~\ref{fig:OFAD} shows some example faces. OFAD is a comprehensively annotated dataset that contains five kinds of occupations: actor, singer, doctor, teacher and farmer. Each occupation includes two range of ages, i.e., 30-50 and 50-80. To the best of our knowledge, it is the first face aging dataset with occupational information. We believe this dataset is benefit for the researches in face progression under different occupations.

We further present a realistic image generation for aging progression under different occupations via the proposed occupational-aware adversarial face aging network which is referred as OAFA. Different from previous approaches which only have one output for one age group, our OAFA can generate several outputs of different occupations.

OAFA has three major components: 1) the generator/encoder aims to generate different future looks for a young face of different occupations, 2) the decoder brings the future looks back to the young face, and 3) the discriminator aims to encourage generator/encoder to generate high quality images of different occupations. The three major components make up two networks: personalized network and occupational-aware adversarial network. The personalized network is an autoencoder network which is formed by the generator/encoder and the decoder. We propose a personalized loss to make the original face can be regenerated by the future looks to keep personalized facial characteristics. The occupational-aware adversarial network consists of the generator/encoder and the discriminator components. The proposed occupational loss includes two terms: conditional adversarial loss~\cite{goodfellow2014generative,mirza2014conditional} and triplet rank loss~\cite{Lai2015Simultaneous,Norouzi2012Hamming}, which aims to obtain the visually plausible texture changes, i.e., skin aging, for different occupations.

Our contributions can be summarized as following:
\begin{itemize}
\item We introduce an occupational face aging dataset which includes several occupations. This helps to explore the effects of the occupations in face age progression problem.

\item We propose an occupational-aware face aging adversarial network to generate multi outputs for occupational-aware age progression problem, which can model the personalities and occupational characteristics of the persons in the face aging process.

\item The empirical results demonstrate the superiority of the proposed method over the state-of-the-arts baseline methods, in which more aging details (textures) and more realistic face images are generated.

\end{itemize}

\section{Related Work}

Many face age progression approaches have been proposed to model the dynamic aging process, which can be mainly divided into two categories, physical model-based~\cite{fu2010age} and prototype-based methods~\cite{tiddeman2001prototyping}. The physical model-based methods~\cite{lanitis2002toward,ramanathan2006modeling,ramanathan2008modeling,suo2012concatenational,suo2010compositional} simulate face aging by modelling the aging mechanisms, e.g., skins, muscles, wrinkle, etc, via employing the parametric model. However, these methods are computationally expensive and require a long age span of each individual. Unfortunately, collecting a wide range of ages of the same person is very difficult or even unlikely, and few of face aging datasets satisfy this requirement.

The prototype-based methods~\cite{burt1995perception,kemelmacher2014illumination} use non-parametric model. They firstly divide faces into groups by age and then the average face is computed for each age group. The average face is referred as prototype and the differences between prototypes are viewed as the aging pattern. The main problem of prototype-based methods is that they may ignore the personalized information, e.g., wrinkles. To persevere the personality, Shu et al.~\cite{shu2015personalized} presented an age progression method based on dictionaries. 
Each group has one dictionary, and two neighbouring groups are linked together for learning the aging pattern. Moreover, the personalized layer is proposed to keep the personalized information.

Deep learning methods have also been proposed for solving the age progression problem. Wang et al.~\cite{wang2016recurrent} proposed a recurrent face aging framework based on a recurrent neural network, which can age the face gradually and keep the personalized information by memorizing the previous faces. Zhang et al.~\cite{zhang2017age} presented a conditional adversarial autoencoder network (CAAE) for learning a face manifold. Their method is based on the conditional generative adversarial networks (CGAN)~\cite{goodfellow2014generative,mirza2014conditional}, which shows impressive results in image generation.

 However, almost all the existing works do not consider that person's appearance may be different under different occupations. To facilitate the researches, we introduce an occupational face aging dataset for exploring the effects of the occupations. We also propose a new occupational-aware adversarial face aging network for age progression. The most similar work to ours is CAAE~\cite{zhang2017age}. Both CAAE and our method utilize autoencoder and CGAN to generate high quality images. The main difference between the CAAE and our method is that we propose personalized network and occupational-aware adversarial network to explicitly pursuit the common age pattern in different ages and occupations. As a result we can obtain more aging details, e.g., the wrinkles and blemishes are more obvious. Different to our method, CAAE assumes the face images lie in a manifold and the autoencoder network is proposed to learn the manifold.



\begin{figure*}[ht]
\begin{center}
\includegraphics[width=1\linewidth]
                   {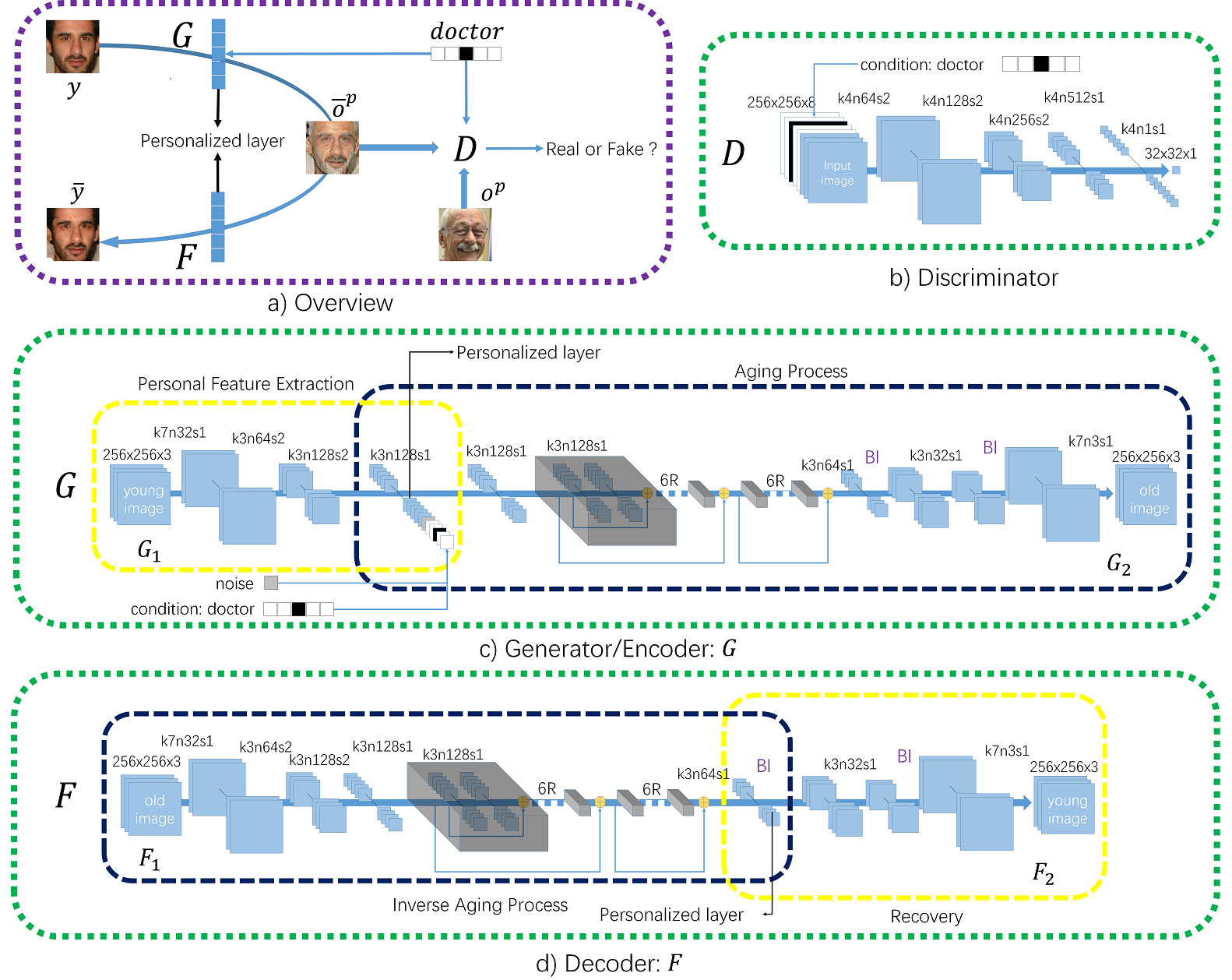}
\end{center}
   \caption{OAFA network for age progression. a) is the overview of OAFA. $G$ is a generative/encoder network that predicts the future look of an input faces $y$ with a certain occupational condition $\ell^p$. Then $F$ maps the generated face $\bar{o}^p$ back to the input face for keeping personality. The discriminator $D$ encourages $G$ to generate old faces that 1) indistinguishable from the real old images and 2) distinguishable from other generated faces of different occupations. The details of architectures of the three networks with corresponding kernel size (k), feature maps (n), stride (s) and residual block (R) are shown in b), c) and d), respectively. }
\label{fig:Generator}
\end{figure*}

\section{The Occupational Face Aging Dataset}

\subsection{Data Statistics}
We collect a dataset of people in 5 different occupations for analysing the occupational effects, which is referred to occupational face aging dataset (OFAD). OFAD consists of over 2,000 diverse face images which are divided into five occupations (actor, singer, doctor, teacher, and farmer). Each occupation contains over 200 male images and 200 female images of different races, and all images have obvious texture information. The age range of these occupational images is between 30 and 80. We divide it into two groups: middle age group (30-50) and old age group (50-80). Some example images are shown in Figure~\ref{fig:OFAD}. We also collect about 200 images of persons without occupational information as input for training, in which the ages of these persons are in the range of 15 to 45.

\subsection{Image Collection}
Image search engines such as Google and Bing are common sources for constructing face aging datasets. In addition to these sources, we also collect face images from two available databases, CACD~\cite{chen2014cross} and FGNET~\cite{lanitis2002toward}.

\textbf{Collecting Images from Image Search Engines.} We download face images from two representative image search engines: Google and Bing, and each of them contains a great number of high-quality face images. In order to collect images with accurately occupational information, we use a combination of descriptive words that contain age information and occupation name as keywords. For example, we use ``retired doctor" as keywords to download doctors' faces for old age group (50-80).

\textbf{Collecting Images from CACD.} The CACD dataset contains more than 160,000 images of 2,000 celebrities from 16 to 62 years-old. According to CACD, most of the celebrities' names are crawled from IMDb.com, which is one of the largest online movie database and contains profiles of millions of movies and celebrities, so we download 30-50 and 50-62 years-old face images as the train set of actor.

\textbf{Collecting Images from FGNET.} To evaluate the performance, faces in FGNET are used as a test set, which contains 1,002 images of 82 people with age range from 0 to 69. It includes the ground truth images for evaluation.

\section{Our Approach}
In this section, we introduce Occupational-aware Adversarial Face Aging network (OAFA), which learns the human aging process under different occupations.

We first introduce some notations. We define $Y = \{ y_i \}_{i=1}^{n_y}$ as the young persons' images. And $M$ denotes as a set of middle-age face images and $O$ is elder face images. The $M$ and $O$ have occupational information. Afterwards, we only discuss how to generate the looks of elder people. The generation for middle-age is similar. Let $O = [O^1,\cdots,O^p,\cdots]$, where $O^p = \{ o^p_j \}_{j=1}^{n^p_o}$  denotes a set of images of the persons who have the $p$-th occupation and $n^p_o$ is the number of images. The data distributions are denoted as $y \sim p_{data}(y)$ and $o^p \sim p_{data}(o^p)$ where $p=1,\cdots,5$ in our paper.

As illustrated in Figure~\ref{fig:Generator}, our architecture contains three components: the generative/encoder network $G$, the decoder network $F$, and the discriminative network $D$. Given a young face image $y$, it goes through the multi convolutional layers and is encoded into high-level feature maps which denote as $G_1(y)$. Then, the feature maps conditioned on certain occupation denote as $(G_1(y),\ell^p)$, where $\ell^p$ is one-hot occupational label for the $p$-th occupation. Finally, these conditioned feature maps are encoded into a future look for a certain occupation, $G_2(G_1(y),\ell^p)$. Note that only changing $\ell^p$, we can generate multiple outputs for different occupations. For ease of representation, we define $\bar{o}^p=G(y,\ell^p) = G_2(G_1(y),\ell^p)$. In addition, we have a adversarial discriminator $D$ which aims to distinguish the generated images $\bar{o}^p$ from the real elder images $o^p$. And a decode function $F$ which is to reconstruct its own input, i.e., $y=F(G(y,\ell^p))$.

\subsection{Loss Function}

Our objective contains two type of terms:
1) personalized loss for keeping human identity information and 2) occupational-aware adversarial loss for obtaining the skin changes in different occupations.

\subsubsection{Personalized Loss}
The primary principle of face age progression is to preserve the personality of the input faces. For example, given a young face image $y$ and the generated old face image with the $p$-occupation $\bar{o}^p$, the generated face image and young face image should be recognized as the same person.

To achieve this goal, we utilize the autoencoder approach~\cite{vincent2010stacked,hinton2006reducing}. It includes an encoder and a decoder, in which the encoder learns a representation for an input data and the decoder reconstructs the representation back to its own input. We require that the generated face image should be able to be reconstructed back to the original image as the CycleGAN~\cite{zhu2017unpaired}, i.e., $y \to \bar{o}^p \to \bar{y}$. With this, the generated face image is one of the representation of the input young face image, which helps preserve the features of young image and keep the personality of the young face. Thus, the personalized loss can be formulated as
\begin{equation}
L_{PER}(G,F)=\sum_{p=1}^{P} E_{y{\sim}p_{data}(y)}[||F(G(y,\ell^p)) - y ||_1].
\end{equation}

The personalized loss limits the space of possible mapping function $G$ because the generated images should be reconstructed back to the original images. Hence, the generated images can't be far away from the source domain.

\subsubsection{Occupational-aware Adversarial Loss}
The second principle is to preserve the common age pattern under different occupations, e.g., the generated face image for farmer
should be recognized as a farmer's face. We propose occupational-aware adversarial loss to address this problem.
\begin{equation}
    L_{OCL}(G,D)=L_{CGAN}(G,D)+L_{TRL}(G).
\label{2}
\end{equation}

Inspired by the impressive results in image generation of the conditional generative adversarial network (CGAN)~\cite{goodfellow2014generative,mirza2014conditional}, we adopt it for our human aging process under different occupations. The objective can be expressed as:
\begin{equation}
\begin{aligned}
    L_{CGAN}(G,D)=&\sum_{p=1}^{P} \Big( E_{(o^p){\sim}p_{data}(o^p)}[\text{log}(D(o^p,\ell^p))] \\
                                 {}&+ E_{(y){\sim}p_{data}(y)}[\text{log}(1-D(G(y,\ell^p),\ell^p))]\\
                                 {}&+ E_{(o^q){\sim}p_{data}(o^q)}[\text{log}(1-D(o^q,\ell^p))] \Big),\\
\label{3}
\end{aligned}
\end{equation}
where $q \in [1,\cdots,P]$ but $q \neq p$. $G$ tries to generate image $G(y,\ell^p)$ that looks similar to images from the $p$-th occupation, and $D$ tries to distinguish the real occupational image $o^p$ and the generated image $G(y,\ell^p)$. $G$ minimizes this objective while $D$ aims to maximize it. Note that we also add $E_{(o^q){\sim}p_{data}(o^q)}[\text{log}(1-D(o^q,\ell^p))]$ to distinguish from other generated faces of different occupations.

To further make the generated images of different occupations look different to each other, we add a triplet rank loss~\cite{Lai2015Simultaneous,Norouzi2012Hamming} that is defined as
\begin{equation}
\begin{aligned}
  & L_{TRL}(G)= \sum_{p=1}^{P} \ \max \Big( 0, \epsilon \\ {}& + || E_{(o^p){\sim}p_{data}(o^p)}[o^p] - E_{(y){\sim}p_{data}(y)}[G(y,\ell^p)] \ ||_1  \\
  & - ||E_{(o^q){\sim}p_{data}(o^q)}[o^q] - E_{(y){\sim}p_{data}(y)}[G(y,\ell^p)] \ ||_1 \Big),&\\
\label{4}
\end{aligned}
\end{equation}
where $\epsilon$ is the margin. We explicitly require that the $\bar{o}^p$ should be closer to the images of the $p$-th occupation than other occupations. It is to make the generate image of $p$-th occupation looks similar to the target domain and help to distinguish the multiple output images.

\subsubsection{Full Objective}
Our full objective is
\begin{equation}
\begin{aligned}
  L(G,F,D)=  &\ \lambda L_{PER}(G,F) \\
     {}& + \mu L_{CGAN}(G,D) \\
     {}& + \nu L_{TRL}(G), \\
\label{5}
\end{aligned}
\end{equation}
where $\lambda$, $\mu$, and $\nu$ control the importance of the three objectives. The final optimization problem can be formulated as
\begin{equation}
\min_{G,F} \max_{D} L(G,F,D).
\label{6}
\end{equation}

\subsection{Network Architecture}
\textbf{Generator/Encoder $G$.} Our generator network follows the architectural proposed by Johnson et al.~\cite{johnson2016perceptual} and SRGAN~\cite{ledig2016photo}. It consists of two parts, $G_1$ and $G_2$. $G_1$ is a small network with three convolutional layers, which learns the feature maps that facilitate the following image generation. The kernel size of the first convolutional layer is $7 \times 7$ and the last two conovlutional layers with $3 \times 3$ filter kerners. Each convolutional layer is followed by one instance-normalization layer~\cite{Ulyanov2016Instance} and one ReLU layer. We use two strides in the last two convolutional layers which makes the size of the output be half of the input. Given a input image $y \in \mathbb{R}^{256 \times 256 \times 3}$, the final output of $G_1$ network is $G_1(y) \in \mathbb{R}^{64 \times 64 \times 512}$. The occupational information is one-hot vector, e.g., $[1,0,0,0,0]$ indicates for actor. We resize the one-hot vector into a cube, e.g., $ \ell^p \in \mathbb{R}^{64 \times 64 \times 5}$ where the values in the $p$-th channel are all one, and other channels are zero. Then it is concatenated to the output of $G_1(y)$ and used as the condition.

Given the high-level feature maps $G_1(y)$ and occupational label $\ell^p$ as input, $G_2$ is a residual network~\cite{gross2016training} to generate the realistic face image. Residual connections are powerful method~\cite{he2016deep}, which make the very deep network easily to be trained. Followed the design of~\cite{gross2016training,johnson2016perceptual}, each residual network consists of two convolutional layers with $3 \times 3$ filter kernels and 128 feature maps, each convolutional layer followed by one instance-normalization layers and one ReLU activation function. There are 12 residual networks in total.

After the residual network, we use a bilinear interpolation method~\cite{gribbon2004novel} to upsample the input instead of deconvoluions, since deconvoluions tend to introduce characteristic artifacts~\cite{odena2016deconvolution,chen2017photographic}. Bilinear interpolation is one of the basic resampling techniques and used to produce a reasonably realistic image. Two bilinear interpolations are used to increase the size of the feature maps, each bilinear interpolation followed by a $3 \times 3$ convolutional layer, one instance-normalization layer and one ReLU layer. The last convolutional layer with $7 \times 7$ kernel size and followed by one instance-normalization layer and one \emph{Tanh} layer.

\textbf{Decoder $F$.} The decoder is an inverse generator. The only difference is that we remove the occupational conditions.

\textbf{Discriminator $D$.} Our discriminator adopt $70 \times 70$ PatchGAN~\cite{isola2016image} as our basic framework. the input of $D$ consists of an old image and an condition vector. All LReLU are leaky with slop 0.2.

\section{Experiments}
In the section, we compare the proposed OAFA against several baselines both qualitatively and quantitatively.

\subsection{Implementation of OAFA}

As with previous works, we normalize the value of each pixel of the input images into $(-1, 1)$, because the normalized pixel value will make the training easier and achieve faster convergence. Similarly, when we concat one-hot vector, we also put its values in the specification between -1 to 1. The value of 0 in the one-hot vector corresponds to -1 and the value of 1 corresponds to 1. The output of the proposed architecture is also in range $(-1, 1)$ by using of the \emph{Tanh} layer.

In training, the hyper-parameters are set as $\lambda=10$, $\mu=1$, $\nu=0.1$. The three networks are updated alternatively with a mini-batch size of 1 through the stochastic gradient solver, i.e., ADAM~\cite{kingma2014adam} ($\alpha=0.0002$, $\beta_{1}=0.5$). After nearly 200 epochs, high-quality results can be obtained. During testing, only $G$ is active. Give a young face and a certain occupational condition, $G$ will generate the corresponding aging face.

Generally speaking, training GANs is a difficult issue in practice because of the instability of GANs learning~\cite{Radford2015Unsupervised}. Least square loss performs more stably during training~\cite{Mao2016Least}, and we replace the negative log likelihood objective in Eq~\ref{3}.

\begin{figure}[htb]
\begin{center}
   \includegraphics[width=1\linewidth]{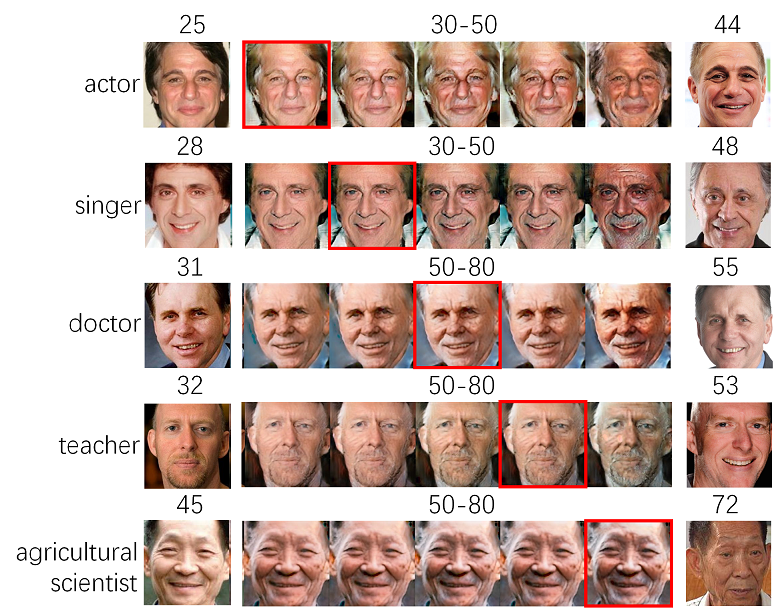}
\end{center}
   \caption{Comparison to the real occupations. The first column is input faces. The second to sixth columns are our results for five occupations (It's hard to find a same farmer's young and old face, we thus replace it with an agricultural scientist). The last column is the ground truth faces. }
\label{fig:compare_occupation}
\end{figure}

\begin{figure}[htb]
\begin{center}
   \includegraphics[width=0.85\linewidth]{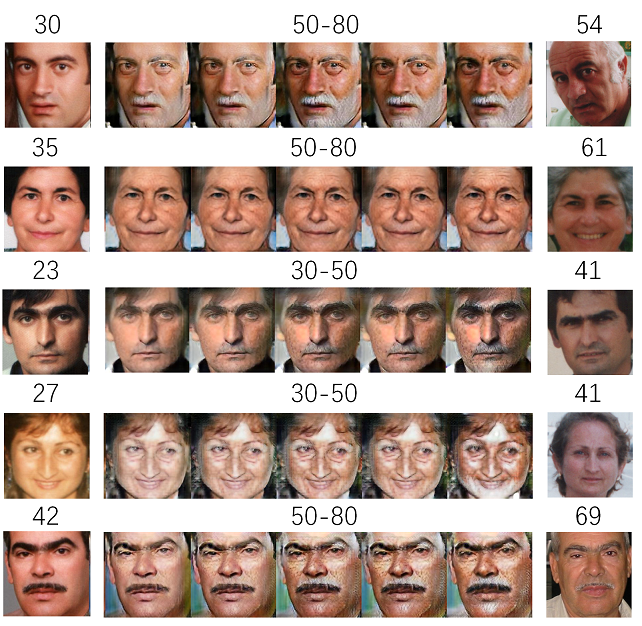}
\end{center}
   \caption{Comparison to the ground truth. The first column is input faces. The second to sixth columns are our results for five occupations (acotr, singer, doctor, teacher, and farmer). The last column is the ground truth faces. Obviously, our method show good performance on details of aging (facial wrinkles, white hair or beard, and skin color) and can preserve personal characteristics.}
\label{fig:compare_FGNET}
\end{figure}

\begin{figure}[htb]
\begin{center}
   \includegraphics[width=1\linewidth]{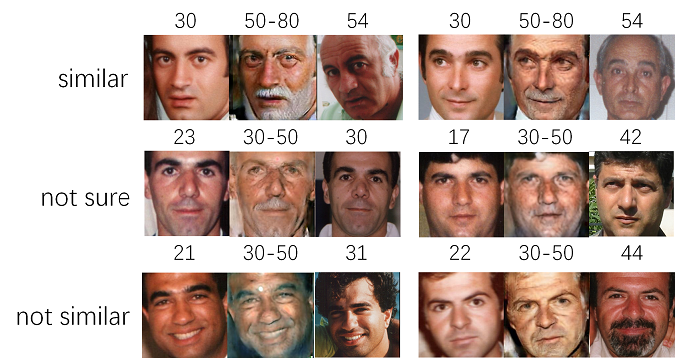}
\end{center}
   \caption{Examples of the quantitative comparison to the ground truth. The two images are indicated as the same person in the first row. The second row is not sure, and the third row is not similar.}
\label{fig:quantitative_groundtruth}
\end{figure}

\begin{figure*}[htb]
\begin{center}
\includegraphics[width=0.7\linewidth]
                   {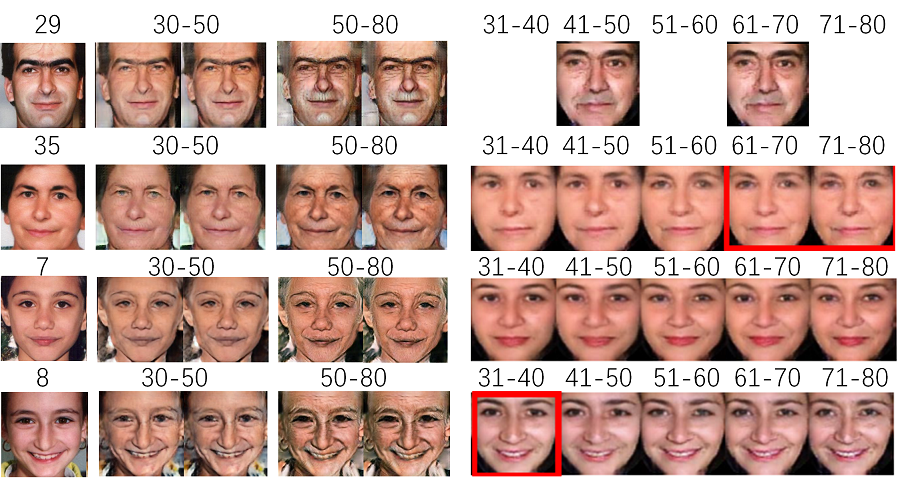}
\end{center}
   \caption{Comparision to CAAE. The first column is input faces. The second to fifth columns are our results for (30-50) and (50-80) age groups, each age group has two pictures, which are teachers and doctors, because the two occupations have the aging characteristics of ordinary people. The sixth to ninth column is the results of CAAE (We cite these picture from the original paper directly, please ignore the red boxes). We can see that the generated faces of CAAE look smooth and young even for 71-80 years-old while the details of skin aging can be clearly seen from our generated images.}
\label{fig:compare_CAAE}
\end{figure*}

\begin{figure}[htb]
\begin{center}
   \includegraphics[width=0.9\linewidth]{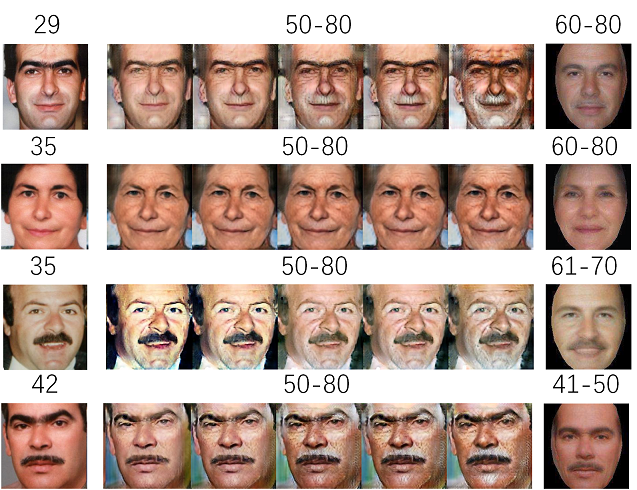}
\end{center}
   \caption{Comparison to RFA. The first column is input faces. The second to sixth columns are our results for five occupations (acotr, singer, doctor, teacher, and farmer). The last column is the result of RFA.}
\label{fig:compare_RFA}
\end{figure}

\begin{figure}[htb]
\begin{center}
   \includegraphics[width=1\linewidth]{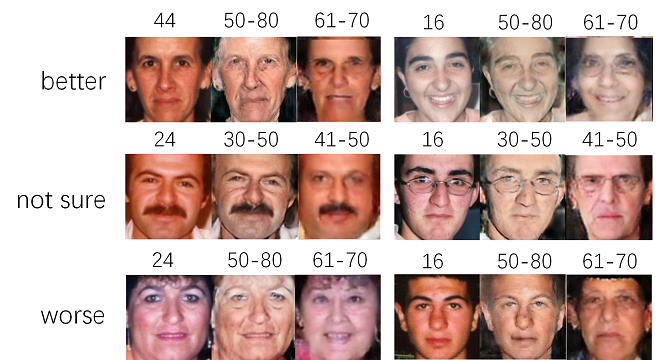}
\end{center}
   \caption{Examples of the quantitative comparison to the prior work. The first row indicates our method is better than prior works. The second row is not sure, and third row is worse.}
\label{fig:quantitative_priorwork}
\end{figure}

\begin{figure}[htb]
\begin{center}
   \includegraphics[width=1\linewidth]{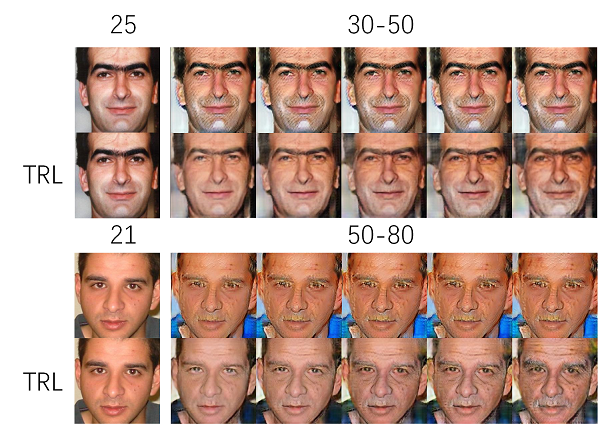}
\end{center}
   \caption{Comparison results of using and not using triplet rank loss in occupational-aware adversarial loss. Without triplet loss, the occupational information can't be obtain accurately.}
\label{fig:loss_relationship}
\end{figure}

\begin{figure}[htb]
\begin{center}
   \includegraphics[width=1\linewidth]{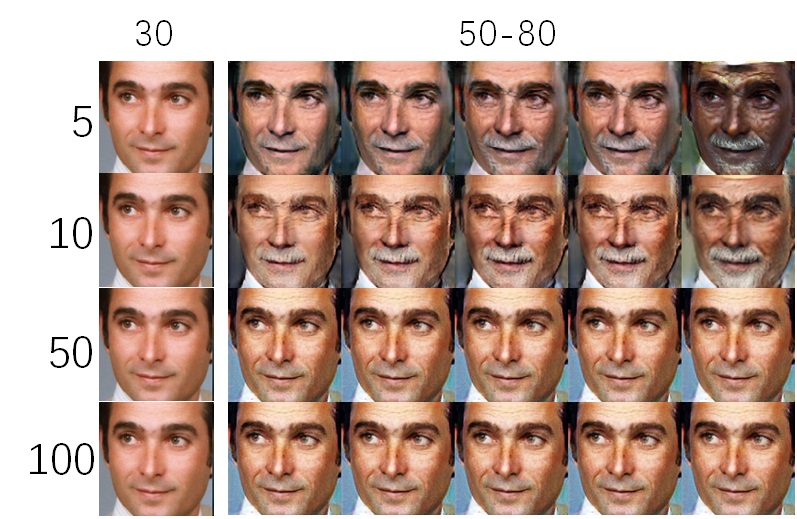}
\end{center}
   \caption{Effects of the personalized loss. As the coefficients $\lambda$ increase, the aging effect and occupation information are less obvious.}
\label{fig:parameter}
\end{figure}

\subsection{Qualitative and Quantitative Comparison}
In this subsection, we evaluate the performance of the proposed method. Following the CAAE~\cite{zhang2017age}, we qualitatively and quantitatively compare with ground truth and the best results from prior works~\cite{zhang2017age,wang2016recurrent}.

\subsubsection{Comparison with Ground Truth}

To qualitatively evaluate the performance, we compare our results with the real face images with occupational information in Figure~\ref{fig:compare_occupation}. We also compare the face images in FGNET~\cite{lanitis2002toward,cootes2008fg} in Figure~\ref{fig:compare_FGNET}. We can see that OAFA can well obtain different textures of different occupations, e.g. wrinkles, hair, blemishes, etc.

For the comparison of quantity, we found 100 volunteers to do the test. Each participant was shown a sequence of  paired images: the original image, our generated image (we randomly selected one in five results), the ground truth. Participants were asked whether the last two images are the same person or not sure. There are 240 paired images of 48 subjects from FGNET in total. First, we randomly selected 40 pairs of images to let participants understand the test process. Then we randomly selected 100 paired images from the rest of the images for testing. 95 valid test results are received, with 60.32\% indicating that the generated face image is the same person as the ground truth, 30.57\% indicating they are not, and 9.11\% not sure. Some test results are shown in Figure~\ref{fig:quantitative_groundtruth}. Qualitative and quantitative comparisons show that our method can obtain realistic images.

\subsubsection{Comparison with State-of-the-arts}

We select CAAE~\cite{zhang2017age} and RFA~\cite{wang2016recurrent} as our baselines. For fair comparison, we use the same input without preprocess to generate images and all results of baselines are directly cited from their original papers. Figure~\ref{fig:compare_CAAE} and Figure~\ref{fig:compare_RFA} show the comparison results. We can see that our method can generate more realist, older images and obtain more skin textures. For example, even 80-years old, the generated faces of CAAE look very young. 
However, the details of skin aging can be clearly seen in the faces generated by our method.

For the comparison of quantity, we also found 100 volunteers to do the test. Each participant was shown a sequence of paired images, the original image, our generated images (we also randomly selected one in five results), and the image generated by prior work (we use the code~\footnote{https://github.com/ZZUTK/Face-Aging-CAAE} of CAAE to generate these image), and asked them which method perform better or not sure. The test set consists of 740 paired images of 148 subjects from FGNET, and some examples were shown in Figure~\ref{fig:quantitative_priorwork}. We randomly selected 40 paired images to let participants understand the test process. Then from the rest of the images, we randomly selected 100 paired images for testing. 92 valid test results are received, with 61.35\% indicating that our method is better, 13.06\% indicating our method is worse, and 25.59\% not sure.

Note that there is no any pre-process to our input images.

\subsection{Effect of the Occupational-aware Adversarial Loss}

For exploring the effects of triplet rank loss, we compare some examples that using and not using the triplet rank loss, i.e., with and without triplet rank loss. The other two losses are fixed. Figure~\ref{fig:loss_relationship} shows the results. We can observe that using the triplet rank loss gives better results.

\subsection{Effect of the Personalized Loss}

For exploring the effects of personalized loss, we also show some exampled results that using different values of $\lambda$ . Figure~\ref{fig:parameter} shows the results which the $\lambda$ is in the range of $[5,10,50,100]$. We can see the generated face images look more like the input young faces with larger weight of personalized loss.

\section{Conclusion}
In this paper, we proposed an occupation-aware face aging progression method via the conditional generative adversarial network. In the proposed deep adversarial architecture, an input young face image conditioned on an occupation goes through the generator network and the future look is generated. Then, we present personalized network and  occupational-aware adversarial network for preserving personality and generate more realistic images for skin changes , respectively. Empirical evaluations on both the qualitative and quantitative comparisons demonstrate the appealing performance of our method.

In future work, we plan to study shape change, e.g., child growth, to obtain more realistic face images.

{\small
\bibliographystyle{ieee}
\bibliography{egbib}
}

\end{document}